\documentclass[twoside,leqno,twocolumn]{article}
\usepackage[letterpaper]{geometry}

\usepackage{ltexpprt}

\usepackage{amsmath,amsfonts,bm}

\def\eqref#1{equation~\ref{#1}}

\def\1{\bm{1}}

\def\rmX{{\mathbf{X}}}

\def\vh{{\bm{h}}}

\def\vm{{\bm{m}}}

\def\mX{{\bm{X}}}

\DeclareMathAlphabet{\mathsfit}{\encodingdefault}{\sfdefault}{m}{sl}
\SetMathAlphabet{\mathsfit}{bold}{\encodingdefault}{\sfdefault}{bx}{n}

\def\gE{{\mathcal{E}}}

\def\gG{{\mathcal{G}}}

\def\gM{{\mathcal{M}}}
\def\gN{{\mathcal{N}}}

\def\gR{{\mathcal{R}}}

\def\gV{{\mathcal{V}}}

\def\gY{{\mathcal{Y}}}

\def\sR{{\mathbb{R}}}

\newcommand{\R}{\mathbb{R}}

\usepackage{amsthm, amsmath, amssymb}
\usepackage{hyperref}
\usepackage{graphicx}
\usepackage{url}

\usepackage[caption=false,font=small]{subfig}
\usepackage{algorithm}
\usepackage{algorithmic}
\usepackage{multirow}
\usepackage{xcolor}

\theoremstyle{definition}

\newcommand{\name}{SparGCP}

\begin{document} 

\title{Graph Sparsification for Enhanced Conformal Prediction in Graph Neural Networks}


\author{
  Yuntian He\\
  The Ohio State University\\
  \texttt{he.1773@osu.edu} \\
  \and
  Pranav Maneriker \thanks{This work was done when the author was a student at The Ohio State University.}\\
  The Ohio State University\\
  \texttt{maneriker.1@osu.edu} \\
  \and
  Anutam Srinivasan\\
  The Ohio State University\\
  \texttt{srinivasan.268@osu.edu} \\
  \and
  Aditya T. Vadlamani\\
  The Ohio State University\\
  \texttt{vadlamani.12@osu.edu} \\
  \and
  Srinivasan Parthasarathy\\
  The Ohio State University\\
  \texttt{srini@cse.ohio-state.edu} \\
}
\date{}

%

\newcommand{\fix}{\marginpar{FIX}}
\newcommand{\new}{\marginpar{NEW}}



\maketitle

\begin{abstract}
\small\baselineskip=9pt Conformal Prediction is a robust framework that ensures reliable coverage across machine learning tasks. Although recent studies have applied conformal prediction to graph neural networks, they have largely emphasized post-hoc prediction set generation. Improving conformal prediction during the training stage remains unaddressed. In this work, we tackle this challenge from a denoising perspective by introducing \name, which incorporates graph sparsification and a conformal prediction-specific objective into GNN training. \name~employs a parameterized graph sparsification module to filter out task-irrelevant edges, thereby improving conformal prediction efficiency. Extensive experiments on real-world graph datasets demonstrate that \name~outperforms existing methods, reducing prediction set sizes by an average of 32\% and scaling seamlessly to large networks on commodity GPUs.
\end{abstract}

\section{Introduction}

Machine learning has been widely adopted across various domains, from healthcare and drug discovery to social networks and recommendation systems \cite{panesar2019machine, gurukar2022multibisage, kang2018self}. Its remarkable performance has driven substantial investment in AI systems. However, a key obstacle to broader applications is the lack of robust performance guarantees. Research in machine learning has traditionally focused on generating accurate predictions, but many often need help to quantify the uncertainty of their outcomes, which hinders the reliable use of machine learning. Conformal Prediction (CP) \cite{vovk2005algorithmic} comes into play to address this. CP is a general framework that provides guaranteed marginal coverage. Specifically, given a trained machine learning model and a desired miscoverage rate $\alpha \in (0, 1)$, CP learns the predicted probabilities on a calibration dataset and then constructs a prediction set (or an interval in regression tasks) for a test sample, ensuring that the ground truth outcome is included with a probability of at least $1 - \alpha$. While maintaining the coverage, CP should also be efficient with respect to the size (or length) of its predictions. Recently, CP has been utilized in diverse AI-based applications to enhance model robustness, including image classification \cite{angelopoulos2021uncertainty}, sequence recommendation \cite{wang2024confidence}, causal inference \cite{lei2021conformal}, and question answering \cite{fisch2021efficient}.

Conformal prediction has been applied to machine learning tasks across different data formats, including graphs. Graphs are powerful data structures that capture complex relationships between distinct entities. Graph-based learning models, such as graph neural networks (GNNs) \cite{kipf2016variational, kipf2017semi}, have successfully demonstrated excellent performance in tasks such as node classification and link prediction. Zargarbashi et al. \cite{zargarbashi2023conformal} and Huang et al. \cite{huang2023uncertainty} paved the way for extending conformal prediction to graph learning by first presenting theoretical insights into the validity of \textit{exchangeability} in transductive node classification. In addition to integrating existing CP methods such as APS \cite{romano2020classification} and TPS \cite{sadinle2019least} into GNNs, recent studies have sought to develop CP approaches tailored explicitly for graph learning tasks \cite{huang2023uncertainty, zargarbashi2023conformal, clarkson2023distribution, zhao2024conformalized, davis2024valid}. However, most of these techniques focus on the post-hoc procedure of conformal prediction, which wraps around a trained model. The question of enhancing conformal prediction during the training stage of GNNs is largely unexplored.

Another challenge of graph conformal prediction is denoising. Given that graph data is non-i.i.d., graph models are susceptible to noisy graph structure data. GNNs learn representations through the message-passing mechanism along the edges between nodes. Understanding the impact of each edge on performance in a specific downstream task is crucial. Crucially, graph sparsification (GS) can be leveraged to filter out task-irrelevant edges and improve the performance of GNNs. 
Previous works have utilized graph sparsification approaches, including both model-free \cite{salha2019degeneracy, razin2023ability} and trainable techniques \cite{zheng2020neuralsparse, luo2021ptdnet, li2023interpretable}, to enhance the performance of GNNs in traditional machine learning tasks. Accordingly, we are interested in exploring whether graph sparsification can effectively facilitate conformal prediction.

\smallskip
\noindent \textbf{Key Contributions:} In this paper, we study the problem of enhancing graph conformal prediction during the training stage through graph sparsification and propose the framework \textit{\underline{Spar}sified \underline{G}raph \underline{C}onformal \underline{P}rediction} (\name). Specifically, we introduce a parameterized graph sparsification module and a CP-based loss to the base GNN model. The graph sparsification module is designed to assess the contribution of each edge to CP tasks and remove noisy edges with lower scores. The CP-based objective is used to simulate the non-conformity metric used by the CP method, which affects the size of prediction sets. \name~is agnostic to the base model and compatible with any GNN classifier such as GCN \cite{kipf2017semi} and GAT \cite{kang2018self}. By jointly training the backbone GNN and the graph sparsifier, \name~effectively reduces noisy edges and lowers the average size of conformal predictions.
Furthermore, \name~can be trained in mini-batches, ensuring scalability over limited GPU memory. We conduct extensive experiments on real-world graph datasets and compare our work with various baselines. Our results show that \name~outperforms other baselines, achieving an average reduction of $32\%$ in the prediction set size and scaling well on large-scale networks including Ogbn-Products.

\section{Preliminaries and Problem Statement}

We first introduce the notations used throughout this paper. Let $\gG = (\gV, \gE, \mX)$ be an attributed graph, where $\gV$ is the set of nodes and $\gE$ is the set of edges. We denote the number of nodes as $n = |\gV|$ and the number of edges as $m = |\gE|$. $\mX \in \R^{n \times d}$ is the node attribute matrix where $d$ is the dimensionality of node features. Let $\gY$ denote the prediction space for machine learning tasks. In this paper, we focus on the node classification problem, where $\gY$ is a set of discrete labels.

\paragraph{Graph Sparsification} Given a graph $\gG = (\gV, \gE)$, graph sparsification generates a subgraph of $\gG$ constructed by the subsets of the original nodes and edges, i.e., $\gG' = (\gV', \gE')$ such that $\gV' \subseteq \gV$ and $\gE' \subseteq \gE$. GS can be achieved either through heuristic algorithms \cite{seidman1983kcore, Rong2020DropEdge, razin2023ability} or by training models to preserve performance in specific tasks \cite{zheng2020neuralsparse, li2023interpretable}.

\paragraph{Graph Neural Networks} GNNs embed both graph structure and node attribute information into vertex representations \cite{kipf2017semi, velickovic2018graph}. Generally, a GNN consists of $l$ layers. Let $\vh_{0}(u) = \mX_{u,:}$ be the attribute vector of node $u$ as its input in the first layer. The $i$-th GNN layer consists of three phases: (1) \textit{Message passing}, which encodes the message $\vm_{i}(u, v) = \mathrm{MP}(\vh_{i-1}(u), \vh_{i-1}(v))$ along the edge between nodes $u$ and $v$. (2) \textit{Information aggregation}, which computes the intermediate representation of a node by aggregating the messages from its neighbors as $\bar{\vm}_{i}(u) = \mathrm{AGGR}(\{\vm_{i}(u, v) \mid v \in \gN(u)\})$, where $\gN(u)$ denotes the neighbors of $u$. (3) \textit{Update}, which applies a non-linear activation $\vh_{i}(u) = \sigma (\vh_{i-1}(u), \bar{\vm}_{i}(u))$ to generate the hidden representation of the current layer. The $l$-th GNN layer outputs $\vh_{l}(u)$ as the final representation for node $u$. For node classification, GNNs are usually followed by a classifier and jointly trained in a supervised manner to predict a label $y \in \gY$ based on the learned node representations.

\begin{figure*}[!t]
\begin{center}
\begin{minipage}[c]{.72\linewidth}
  \includegraphics[width=\linewidth]{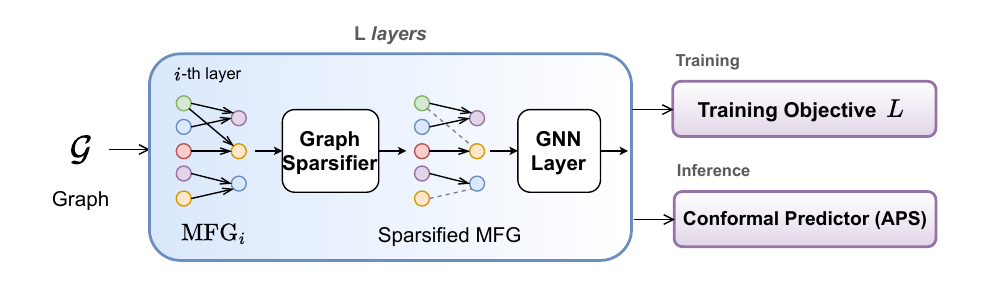}
  \caption{Model Architecture.}
  \label{fig:model_arch}
\end{minipage}\hfill
\begin{minipage}[c]{.24\linewidth}
  \includegraphics[width=\linewidth]{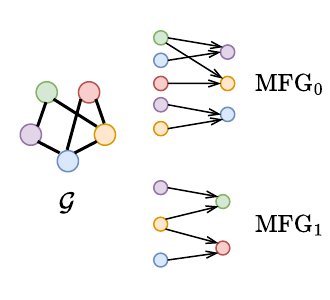}
  \caption{MFG example.}
  \label{fig:mfg_example}
\end{minipage}%
\end{center}
\end{figure*}

\paragraph{Conformal Prediction} Conformal node classification aims to train a classifier $f: \rmX \rightarrow 2^{\gY}$ that outputs a prediction set that covers the ground truth label with a probability of at least $1 - \alpha$. Our study focuses on the transductive and split conformal prediction setting \cite{papadopoulos2002inductive}, where the graph $\gG$ remains fixed during training, and the calibration process in CP is conducted on a standalone calibration set. Specifically, all labeled nodes $\gV_{L}$ are partitioned into four disjoint subsets $\gV_{\mathrm{train}}, \gV_{\mathrm{valid}}, \gV_{\mathrm{calib}}, \gV_{\mathrm{test}}$. In the training stage, the entire graph structure, node attribute $\mX$, and labels of $\gV_{\mathrm{train}}, \gV_{\mathrm{valid}}$ are visible to the model, while the labels of $\gV_{\mathrm{calib}}, \gV_{\mathrm{test}}$ are only revealed in the calibration and testing phases. 

Generally, CP methods construct the prediction set using a given trained classifier in three steps. (1) It computes the non-conformity scores over $\gV_{\mathrm{calib}}$. Considering the predicted probability distribution of node $u$ as $\hat{p}(y\mid u)$ for $\forall y \in \gY$ and its ground truth label $y_{u}$, it defines a non-conformity score denoted as $V(u, y_{u})$. (2) Then it calculates the $(1-\alpha)(1 + \frac{1}{n})$ quantile of the non-conformity scores of $\gV_{\mathrm{calib}}$ as the prediction threshold $\hat{\eta}$. (3) Finally, for a test sample $v \in \gV_{\mathrm{test}}$, the prediction set includes every label with a non-conformity score lower than $\hat{\eta}$. Note that the coverage guarantee of graph conformal prediction relies on the assumption of exchangeability between nodes in $\gV_{\mathrm{calib}} \cup \gV_{\mathrm{test}}$, which has been shown to hold by previous studies \cite{huang2023uncertainty, zargarbashi2023conformal}.

A key metric of CP methods is \textit{efficiency}, which is measured by the average size of their prediction sets. Intuitively, a smaller prediction set size indicates that the CP approach is capable of generating accurate and concise predictions.

\paragraph{Problem Statement} Given an attributed graph $\gG = (\gV, \gE, \mX)$ and the prediction space $\gY$, we aim to leverage graph sparsification techniques to enhance the training of a conformal GNN classifier $f: \gV \rightarrow 2^{\gY}$. The GNN trained on the sparsified graph $\gG'$ should effectively generalize over $\gG$ while minimizing $\frac{1}{|\gV_{\mathrm{test}}|}\sum_{u \in \gV}|f(u)|$ to deliver efficient conformal prediction performance.

\section{Model Architecture}

In this section, we propose the graph conformal prediction framework \name. The key idea is to leverage layer-wise graph sparsification to filter out noisy messages in the graphs and train the model with a CP-based objective to improve prediction efficiency. Moreover, \name~supports mini-batch training to ensure scalability. Figure \ref{fig:model_arch} presents the overview of \name.

\paragraph{Layer-wise Graph Sparsification}

Earlier works on graph sparsification \cite{seidman1983kcore, spielman2011spectral, batson2009twice} primarily focused on finding subgraphs that preserve structural properties in an unsupervised manner. More recent studies have shifted towards parameterizing the sparsification procedure by developing specific GS modules jointly trained with GNN backbones \cite{zheng2020neuralsparse, li2023interpretable, liu2023comprehensive}. Generally, these approaches utilize node or edge features to evaluate edge importance and remove those that are less informative. We adopt this idea to effectively integrate supervision signals and enhance the prediction efficiency.

Specifically, \name~adds a graph sparsifier prior to each GNN layer, as illustrated in Algorithm \ref{alg:gs}. The input of our sparsification algorithm is a \textit{message flow graph} (MFG), which simulates the message passing procedure for the current layer. The MFG for $i$-th GNN layer is a directed bipartite graph, where nodes at layer $i$ (denoted as $\gV_{i, R}$) receive necessary inputs from the previous layer (denoted as $\gV_{i, L}$). We iteratively construct MFGs from the last layer to the first layer. Each MFG is typically generated by performing a breadth-first search or sampling from the given seed nodes. Figure \ref{fig:mfg_example} presents a toy example of MFGs for a 2-layer GNN with a batch size of 2, where the green and red nodes on the right side of $\mathrm{MFG}_{1}$ are the seed nodes of this batch.

For each edge in the MFG, \name~employs a scoring module to assess its importance to the downstream CP task. In this work, we use a two-layer MLP as our scoring module. Then it computes the $1-\gamma$ quantile of the edge scores, denoted as $\hat{\eta}_{e}$, where $\gamma$ is a hyperparameter that controls the proportion of edges that will be filtered out in the MFG. Finally, \name~removes the edges with scores lower than $\hat{\eta}_{e}$. Note that the MFG reduction procedure at the $l$-th layer is equivalent to graph sparsification to the original graph.

A key advantage of using MFGs is scalability. Most existing parameterized GS approaches operate on the entire graph, which can become a bottleneck when dealing with large graph data or limited hardware capacity (e.g., GPU memory). In contrast, MFG supports minibatch training and therefore ensures scalability. Specifically, \name~randomly partitions nodes in the original graph into smaller batches and constructs MFGs with the corresponding seed nodes of each batch. This allows for flexible control of the computational overhead of each batch and enables smooth scaling across different datasets and computational resources.

\begin{algorithm}[!t]
\caption{Graph Sparsification Module}\label{alg:gs}
\begin{algorithmic}[1]
\REQUIRE MFG $\gM_{i} = (\gV_{i, L}, \gV_{i, R}, \gE_{i})$, $\vh_{i-1}$, $\gamma$
\ENSURE Sparsified MFG $\gM'_{i} = (\gV_{i, L}, \gV_{i, R}, \gE'_{i})$ such that $\gE'_{i} \subseteq \gE_{i}$
\FOR{$(u, v) \in \gE_{i}$}
    \STATE $\bm{\pi}_{u, v} \gets \mathrm{Scoring}(\vh_{i-1}(u), \vh_{i-1}(v))$
\ENDFOR
\STATE $\hat{\eta}_{e} \gets \mathrm{Quantile}(1 - \gamma, \{ \bm{\pi}_{u, v} \mid (u, v) \in \gE_{i}\})$
\STATE $\gE'_{i} \gets \{ (u, v) \mid \bm{\pi}_{u, v} \geq \hat{\eta}_{e} \}$
\end{algorithmic}
\end{algorithm}

\begin{table*}[t]
\caption{Statistics of datasets}
\label{tbl:dataset}
\begin{center}
\begin{tabular}{lrrrr}
\multicolumn{1}{c}{\bf Dataset}  &\multicolumn{1}{c}{\# Nodes}  & \multicolumn{1}{c}{\# Edges} &\multicolumn{1}{c}{\# Features}  & \multicolumn{1}{c}{\# Classes}
\\ \hline \\
CiteSeer & 3,327 & 9,228 & 3,703 & 6 \\
Cora & 19,793 & 126,842 & 8,710 & 70 \\
Co-CS & 18,333 & 163,788 & 6,805 & 15 \\
Amzn-Comp & 13,752 & 491,722 & 767 & 10 \\
Ogbn-Arxiv & 169,343 & 1,166,243 & 128 & 40 \\
Ogbn-Products & 2,449,029 & 61,859,140 & 100 & 47 \\
\end{tabular}
\end{center}
\end{table*}

\paragraph{Training Objectives} 

\name~jointly trains the graph sparsifier and GNN layers through two distinct objectives. We first adopt a (semi-)supervised cross-entropy loss, denoted as $L_{ce}$, for node classification. To further optimize conformal prediction during the training stage, we devise a new CP-based loss aimed at enhancing prediction efficiency. Specifically, the proposed objective simulates the non-conformity score proposed by the popular CP method Adaptive Prediction Sets (APS) \cite{romano2020classification}. For each sample $u$ in the calibration set $\gV_{\mathrm{calib}}$, APS ranks the labels in a sequence $\{y_{(1)}, y_{(2)}, \dots, y_{(|\gY|)}\}$ such that $\hat{p}(y_{(i)} \mid u) \geq \hat{p}(y_{(j)} \mid u)$ for all $i < j$. Let $\hat{\phi}(y_{u})$ represent the index of the ground truth label $y_{u}$ in the sorted sequence. Then APS defines the non-conformity score as
\begin{equation}
    V(u, y_{u}) = \sum_{i = 1} ^ {\hat{\phi}(y_{u})} \hat{p}(y_{(i)} \mid u)
\label{eqn:aps_score}
\end{equation}


\name~integrates model training with a supervision signal that aligns with APS. For each training batch $\gV_{\mathrm{b}} \subseteq \gV_{\mathrm{train}}$, \name~first generates predicted probabilities $\{ \hat{p}(y \mid u) \mid \forall y \in \gY, u \in \gV_{b} \}$ and calculates non-conformity scores for all nodes, as defined in Equation \ref{eqn:aps_score}. The CP-based loss is then determined by the estimated prediction threshold over the training batch. Formally, given a hyperparameter $\alpha_{\mathrm{train}}$, the CP-based objective is defined as
\begin{equation}
    L_{cp} = \mathrm{Quantile}(1 - \alpha_{\mathrm{train}}, \{V(u, y_{u}) \mid u \in \gV_{b}\})
\end{equation}
The intuition behind $L_{cp}$ is to use the training data to estimate the prediction threshold, which will be later calculated over $\gV_{\mathrm{calib}}$, and optimize it by adjusting the prediction distribution. When the prediction threshold $\hat{\eta}$ is high, a CP classifier needs to include more labels to meet the coverage requirement, resulting in prediction inefficiency. Therefore, \name~minimizes $L_{cp}$ to encourage smaller prediction sets. Finally, the overall training objective of \name~is formulated as
\begin{equation}
    L = L_{ce} + \lambda L_{cp}
\end{equation}
where $\lambda \in \sR$ is a hyperparameter that controls the strength of the CP-based loss.

\section{Experiments}

\subsection{Experimental Setup}

\paragraph{Datasets:}
We evaluate our work on six datasets for node classification, including citation networks \texttt{CiteSeer} and \texttt{Cora}, academic network \texttt{Co-CS}, purchase graph \texttt{Amzn-Comp}, and popular benchmark datasets \texttt{Ogbn-Arxiv} and \texttt{Ogbn-Products}. All datasets are from Deep Graph Library (DGL) \cite{wang2019dgl} and Open Graph Benchmark (OGB) \cite{hu2020open}. \autoref{tbl:dataset} presents the data statistics for each dataset. See Appendix \ref{apx:datasets} for additional dataset descriptions.

\smallskip
\noindent{\bf Baselines:}
We compare our method with various baselines. First, we use \textit{GCN} \cite{kipf2017semi} and \textit{GAT} \cite{kang2018self} as two vanilla GNN baselines, which also serve as the GNN backbones for other baseline approaches. Additionally, we have (1) \textit{$k$-core} \cite{seidman1983kcore}: A heuristic graph sparsification method that generates a subgraph where each node has at least $k$ neighbors. It has been leveraged by recent studies \cite{salha2019degeneracy, brandeis2020graph} to improve graph learning performance. Specifically, we train GNNs on graphs sparsified by $k$-core. (2) \textit{DropEdge} \cite{Rong2020DropEdge}: A method that randomly drops a proportion $p$ of edges at each GNN layer during training to enhance robustness and generalization. (3) \textit{NeuralSparse} \cite{zheng2020neuralsparse}: A parameterized graph sparsification model that retains $k$ edges for each node based on edge scores learned from node features and raw adjacency weights. (4) \textit{PTDNet} \cite{luo2021ptdnet}: An edge denoising approach that refines the edge weights in the adjacency matrix, providing more flexibility than traditional GS methods and, therefore, greater capacity.

\begin{table*}[t]
\caption{Comparing APS efficiency with miscoverage rate $\alpha=0.1$ of all baselines with GCN backbone. (Best performance in \textbf{bold}; second-best \underline{underlined}; OOM for out-of-memory)}
\label{tbl:overall_GCN}
\begin{center}
\begin{tabular}{lrrrrrr}
\multicolumn{1}{c}{\bf Dataset}  &\multicolumn{1}{c}{GCN}  &\multicolumn{1}{c}{$k$-core} &\multicolumn{1}{c}{DropEdge}  &\multicolumn{1}{c}{PTDNet}  &\multicolumn{1}{c}{NeuralSparse} & \multicolumn{1}{c}{\name}
\\ \hline \\
CiteSeer & 3.62 {\scriptsize$\pm$ 0.23} & \underline{3.57 {\scriptsize$\pm$ 0.26}} & 3.66 {\scriptsize$\pm$ 0.22} & 4.25 {\scriptsize$\pm$ 0.62} & 3.70 {\scriptsize $\pm$ 0.46} & \textbf{2.99 {\scriptsize$\pm$ 0.36}} \\
Cora  & 22.18 {\scriptsize$\pm$ 1.32} & 23.68 {\scriptsize$\pm$ 1.23} & \underline{19.46 {\scriptsize$\pm$ 0.88}} & OOM & OOM & \textbf{13.00 {\scriptsize$\pm$ 0.87}} \\
Co-CS & 9.65 {\scriptsize$\pm$ 2.64} & 8.61 {\scriptsize$\pm$ 0.21} & \underline{7.59 {\scriptsize$\pm$ 0.43}} & OOM & OOM & \textbf{6.29 {\scriptsize $\pm$ 0.79}} \\
Amzn-Comp & 6.34 {\scriptsize$\pm$ 1.07} & 5.15 {\scriptsize$\pm$ 0.93} & 4.89 {\scriptsize$\pm$ 0.94} & \textbf{3.96 {\scriptsize$\pm$ 0.50}} & 5.63 {\scriptsize$\pm$ 0.46} & \underline{4.17 {\scriptsize$\pm$ 0.70}} \\ 
\vspace{-5pt} \\
\hline \\
Avg. Reduce & - & 6.04\% & \underline{13.84\%} & N/A & N/A & \textbf{31.96\%} \\
\end{tabular}
\end{center}
\end{table*}

\begin{table*}[t]
\caption{Comparing APS efficiency with miscoverage rate $\alpha=0.1$ of all baselines with GAT backbone. (Best performance in \textbf{bold}; second-best \underline{underlined}; OOM for out-of-memory)}
\label{tbl:overall_GAT}
\begin{center}
\begin{tabular}{lrrrrrr}
\multicolumn{1}{c}{\bf Dataset}  &\multicolumn{1}{c}{GAT}  &\multicolumn{1}{c}{$k$-core} &\multicolumn{1}{c}{DropEdge}  &\multicolumn{1}{c}{PTDNet}  &\multicolumn{1}{c}{NeuralSparse} &\multicolumn{1}{c}{\name} 
\\ \hline \\
CiteSeer &  \underline{2.84 {\scriptsize $\pm$ 0.17}} & 2.99 {\scriptsize $\pm$ 0.60} & 2.89 {\scriptsize $\pm$ 0.24}& 3.43 {\scriptsize$\pm$ 0.64} & 3.28 {\scriptsize $\pm$ 0.55} & \textbf{2.30 {\scriptsize $\pm$ 0.30}} \\
Cora & 15.91 {\scriptsize $\pm$ 0.94} & \underline{14.88 {\scriptsize $\pm$ 0.88}} & 16.85 {\scriptsize $\pm$ 1.50} & OOM & OOM & \textbf{11.09 {\scriptsize $\pm$ 0.57}} \\
Co-CS &  8.90 {\scriptsize $\pm$ 0.77} & \underline{6.85 {\scriptsize $\pm$ 0.48}} & 8.84 {\scriptsize $\pm$ 1.01} & OOM & OOM & \textbf{5.07 {\scriptsize $\pm$ 0.87}} \\
Amzn-Comp & 2.91 {\scriptsize $\pm$ 0.34} & 2.89 {\scriptsize $\pm$ 0.38} & \underline{2.68 {\scriptsize $\pm$ 0.30}} & OOM & OOM & \textbf{2.58 {\scriptsize $\pm$ 0.27}} \\
\vspace{-5pt} \\
\hline \\
Avg. Reduce & - & \underline{6.23\%} & 0.23\% & N/A & N/A & \textbf{25.92\%} \\
\end{tabular}
\end{center}
\end{table*}

\smallskip
\noindent{\bf Evaluation Metric} We adopt APS \cite{romano2020classification} as the post-hoc CP method for all baselines and evaluate their efficiency by measuring the average size of the prediction sets.

\smallskip
\noindent{\bf Setup:} For each dataset, we create 20 random splits of the nodes into three disjoint sets $\gV_{\mathrm{train}}, \gV_{\mathrm{valid}}, \gV_{\mathrm{calib}~\cup~\mathrm{test}}$. Then we further randomly partition $\gV_{\mathrm{calib}~\cup~\mathrm{test}}$ into $\gV_{\mathrm{calib}}$ and $\gV_{\mathrm{test}}$ 50 times. In other words, we train each baseline with 20 different train/validation splits and report the average results over 1000 runs. Details of model architecture and hyperparameter selection are in Appendix \ref{apx:hyperparameters}. Most of our experiments were run on a commodity NVIDIA P100 GPU (around 800 USD) in the spirit of democratizing access to our models. We purposefully demonstrate our results on relatively inexpensive commodity hardware. For the much larger graph datasets (OGB), we relied on the slightly more expensive but still affordable NVIDIA A100 chipsets (around 8000 USD).

\subsection{Results}
Tables \ref{tbl:overall_GCN}, \ref{tbl:overall_GAT} present the APS efficiency of all baselines using GCN and GAT as the backbone GNN, respectively. The reported values represent the average prediction set size for each method, where lower values indicate better efficiency. 

\textbf{\name~effectively improves CP efficiency.} Table \ref{tbl:overall_GCN} presents the average prediction size of each baseline. Since all baselines satisfy the empirical coverage constraint, we omit the coverage results from the table.  Our results demonstrate that \name~successfully reduces the prediction size of base GNNs across all four datasets. For example, the average prediction size of GCN on Cora is 22.18, while \name~reduces it by 41.39\% to 13.00. Across these four datasets, \name~improves the efficiency of GCN by an average of 31.96\% and that of GAT by 25.92\%. These results demonstrate that our approach effectively works with different GNNs and consistently enhances their prediction efficiency in CP tasks.

\begin{table*}[t]
\caption{Effectiveness of graph sparsification and CP-based loss. (OOM for out-of-memory)}
\label{tbl:effectiveness}
\begin{center}
\begin{tabular}{lrrrr}
\multicolumn{1}{c}{\bf Baseline}  &\multicolumn{1}{c}{CiteSeer}  &\multicolumn{1}{c}{Cora} &\multicolumn{1}{c}{Co-CS}  &\multicolumn{1}{c}{Amzn-Comp}
\\ \hline \\
\name & 2.99 {\scriptsize$\pm$ 0.36} & 13.00 {\scriptsize$\pm$ 0.87} & 6.29 {\scriptsize$\pm$ 0.79} & 4.17 {\scriptsize$\pm$ 0.70} \\
 $-$ CP-based loss {\scriptsize(GS only)} & 3.53 {\scriptsize$\pm$ 0.32} & 18.57 {\scriptsize$\pm$ 1.56} & 7.60 {\scriptsize$\pm$ 0.58} & 5.07 {\scriptsize$\pm$ 1.15}\\
 $-$ GS {\scriptsize(CP-based loss only)} & 3.21 {\scriptsize$\pm$ 0.33} & 14.63 {\scriptsize$\pm$ 1.14} & 6.76 {\scriptsize$\pm$ 0.90} & 4.97 {\scriptsize$\pm$ 0.88} \\
GCN & 3.62 {\scriptsize$\pm$ 0.23} & 22.18 {\scriptsize$\pm$ 1.32} & 9.65 {\scriptsize$\pm$ 2.64} & 6.34 {\scriptsize$\pm$ 1.07} \\\hline \\
PTDNet & 4.25 {\scriptsize$\pm$ 0.62} & OOM & OOM & 3.96 {\scriptsize$\pm$ 0.50} \\
+ CP-based loss & 3.76 {\scriptsize $\pm$ 0.45} & OOM & OOM & 3.44 {\scriptsize $\pm$ 0.42} \\
\hline \\
NeuralSparse & 3.70 {\scriptsize $\pm$ 0.46} & OOM & OOM & 5.63 {\scriptsize$\pm$ 0.46} \\
+ CP-based loss & 3.68 {\scriptsize $\pm$ 0.45} & OOM & OOM & 4.98 {\scriptsize $\pm$ 0.71} \\
\end{tabular}
\end{center}
\end{table*}

\textbf{\name~outperforms other baselines.} Compared to other baselines, \name~yields the most significant improvement in prediction efficiency. We first observe that $k$-core reduces the prediction set size by an average of 6.04\% with GCN and 6.23\% with GAT. At the same time, DropEdge shows inconsistent performance - achieving significant improvements with GCN but only marginal gain with GAT. These results strongly suggest that model-free approaches cannot consistently enhance CP performance. Moreover, parameterized approaches (PTDNet and NeuralSparse) generally fail to generate efficient conformal predictions, except PTDNet with GCN on the Amzn-Comp dataset. This is because these approaches do not incorporate any CP-specific optimization during training. On the other hand, our approach \name~outperforms other baselines in most cases. This highlights the effectiveness of both our graph sparsification module and CP-based objective in enhancing prediction efficiency.

\begin{figure*}[t]
\begin{center}

\includegraphics[width=0.5\linewidth]{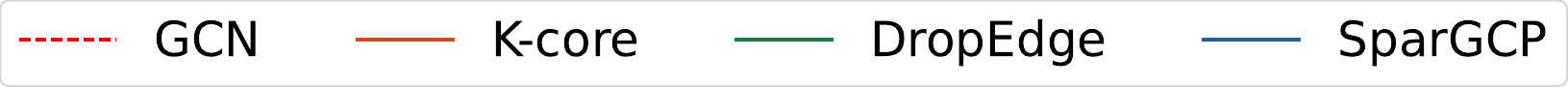}\\

\subfloat[CiteSeer]{\includegraphics[width=0.23\linewidth]{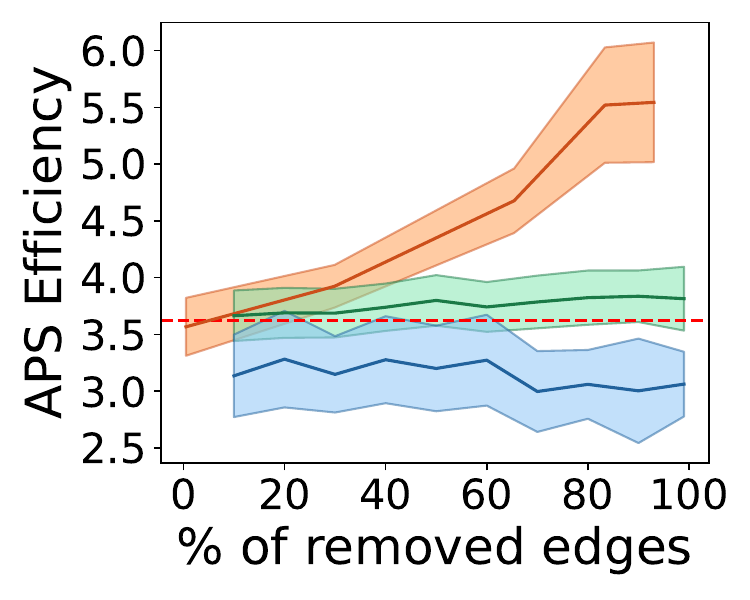}
\label{fig:reducing_edges_GCN_CiteSeer}}
\subfloat[Cora]{\includegraphics[width=0.23\linewidth]{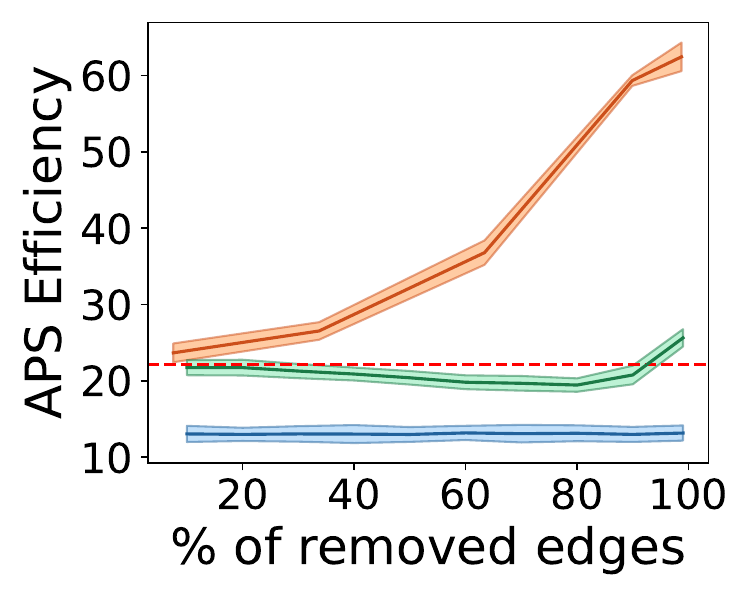}
\label{fig:reducing_edges_GCN_Cora}}
\subfloat[Co-CS]{\includegraphics[width=0.23\linewidth]{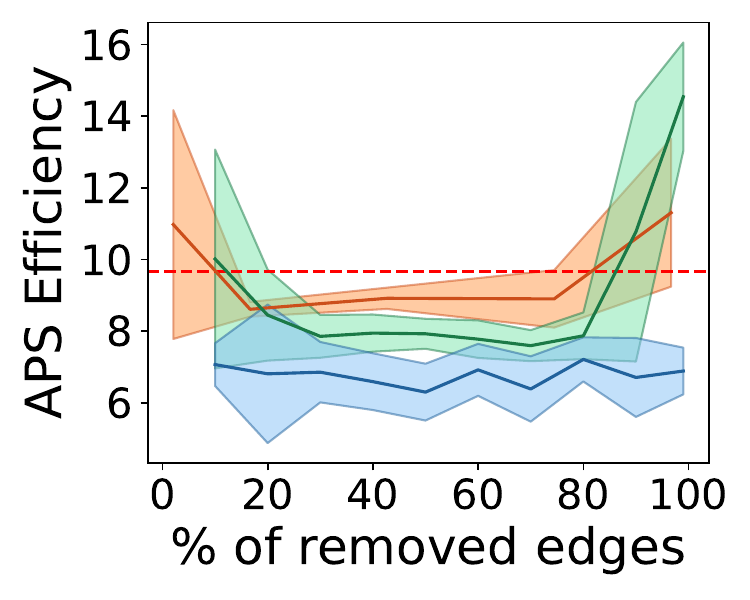}
\label{fig:reducing_edges_GCN_Coauthor_CS}}
\subfloat[Amzn-Comp]{\includegraphics[width=0.23\linewidth]{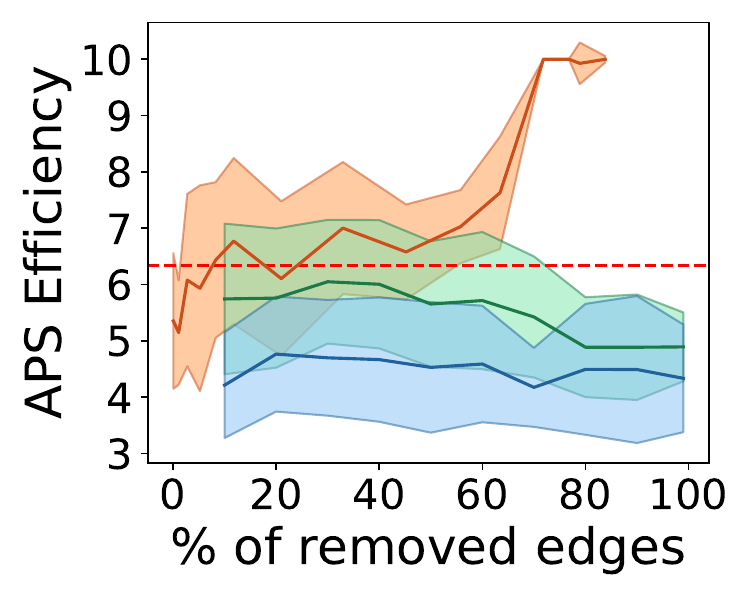}
\label{fig:reducing_edges_GCN_Amazon_Computers}}
\end{center}
\caption{APS efficiency with varying strength of graph sparsification.}
\label{fig:removing_edges}
\end{figure*}

\textbf{\name~is more scalable.} Our design of minibatch training enables scalable training across all datasets. We observe that other parameterized baselines run out of memory on multiple datasets because they require the entire graph structure and full attribute data, which exceeds the memory constraint (16GB on P100). In contrast, \name~can be trained in mini-batches that can efficiently fit within the limitation of GPU memory.

\subsection{Ablation Study}

\paragraph{Effectiveness of GS/CP components.} To investigate the contribution of our graph sparsification module and CP-based loss to the overall performance, we conduct an ablation study by removing each component from our approach. As shown in Table \ref{tbl:effectiveness}, the performance drops after removing either of them, which demonstrates that both designs are critical to CP efficiency. Moreover, we adapt two parameterized baselines by integrating them with the CP-based objective used in our approach. We observe noticeable improvements in their performance, which suggests that our design is also generalizable to other methods.

\paragraph{Impact of sparsification strength on prediction efficiency.} We vary the corresponding hyperparameters to adjust the strength of graph sparsification and observe the impact on the CP efficiency of each baseline. Figure \ref{fig:removing_edges} shows the results of all baselines using GCN as their backbone. Results with GAT are provided in Appendix \ref{apx:sparsification_strength_full}. Generally, model-free baselines ($k$-core and DropEdge) decline significantly when more edges are removed, as these methods cannot differentiate between informative and noisy edges. On the other hand, our work maintains its efficiency when increasing the strength of sparsification and consistently outperforms other baselines.

\begin{figure*}[t]
\begin{center}
\subfloat[CiteSeer]{\includegraphics[width=0.23\linewidth]{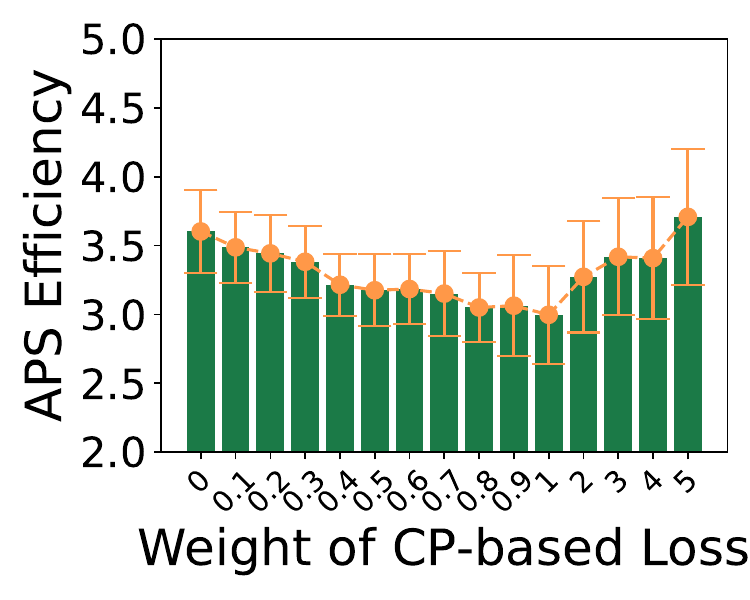}
\label{fig:varying_weight_GCN_CiteSeer}}
\subfloat[Cora]{\includegraphics[width=0.23\linewidth]{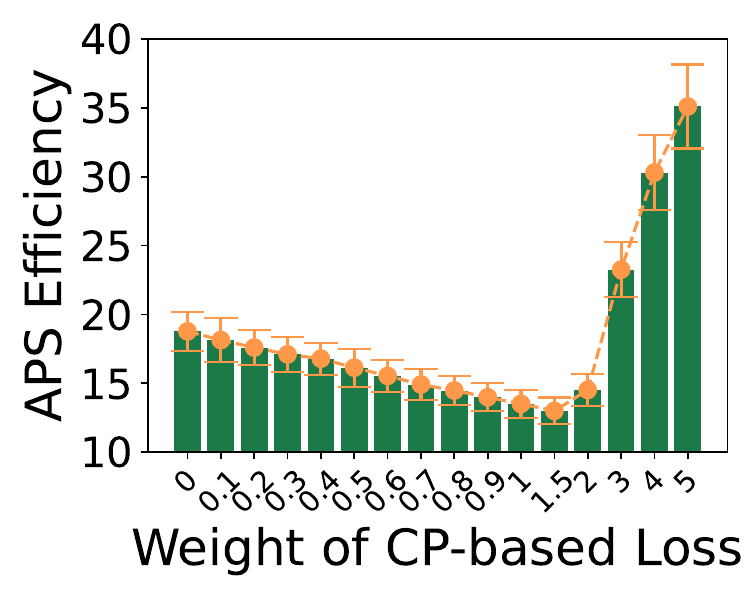}
\label{fig:varying_weight_GCN_Cora}}
\subfloat[Co-CS]{\includegraphics[width=0.23\linewidth]{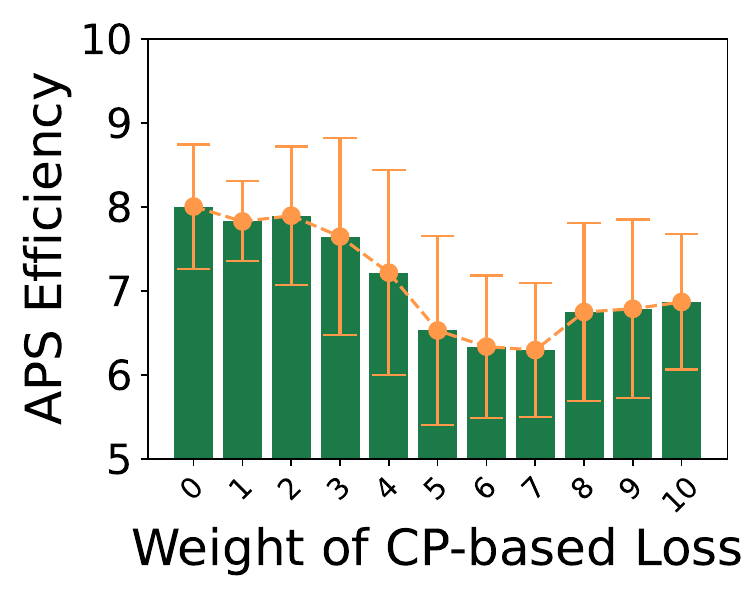}
\label{fig:varying_weight_GCN_Coauthor_CS}}
\subfloat[Amzn-Comp]{\includegraphics[width=0.23\linewidth]{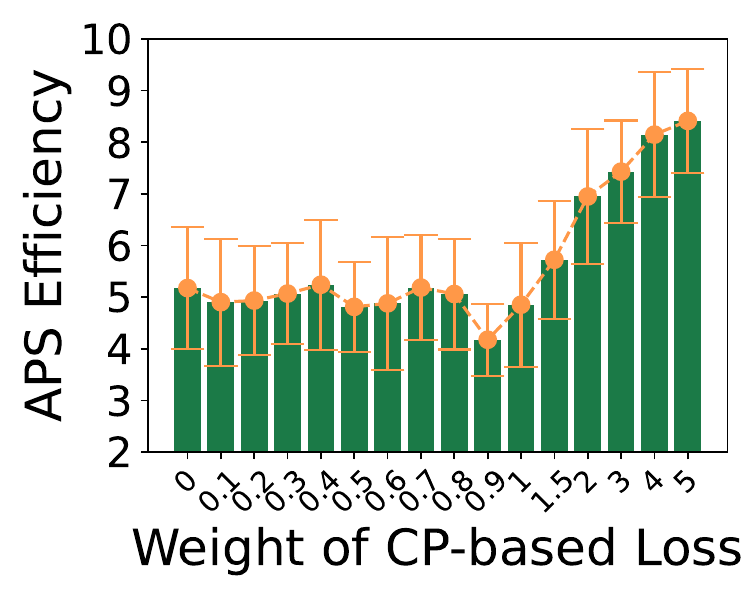}
\label{fig:varying_weight_GCN_Amazon_Computers}}

\end{center}
\caption{Varying the weight of CP-based loss.}
\label{fig:varying_weight}
\end{figure*}

\paragraph{Impact of the weight of CP-based loss.}
We also fix the edge-dropping hyperparameter and vary the weight $\lambda$ of CP-based loss to evaluate its impact on CP efficiency. Figure \ref{fig:varying_weight} presents the result of GCN-based baselines. Results of GAT-based approaches can be found in Appendix \ref{apx:varying_weight_full}. In general, the performance of \name~declines when $\lambda$ is either too small or too large. When $\lambda$ is small, the model is not adequately trained for the CP-based objective. On the other hand, when $\lambda$ is large, the model is overfitted to minimize the quantile of non-conformity scores $\hat{\eta}$ by generating smaller probabilities for ground truth labels, resulting in a lack of generalizability. 

\begin{figure*}[t]
\begin{center}

\includegraphics[width=0.9\linewidth]{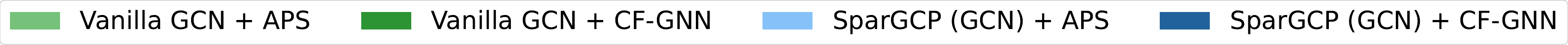}\\

\subfloat[CiteSeer (Efficiency)]{\includegraphics[width=0.23\linewidth]{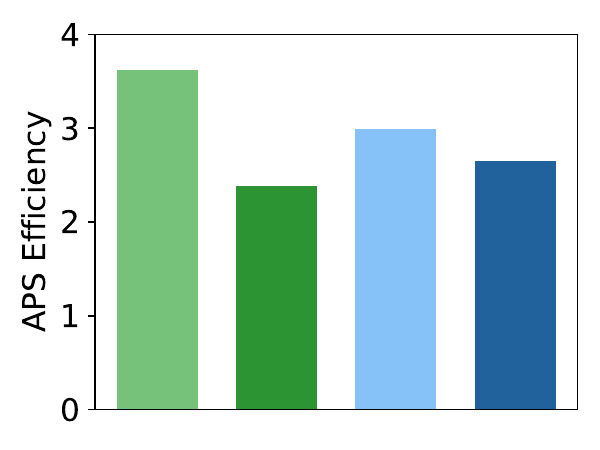}
\label{fig:cfgnn_CiteSeer-GCN_efficiency}}
\subfloat[Amzn-Comp (Efficiency)]{\includegraphics[width=0.23\linewidth]{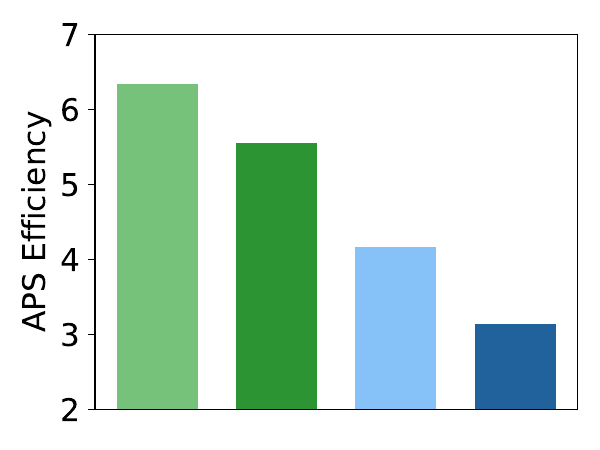}
\label{fig:cfgnn_Ammazon_Computers-GCN_efficiency}}
\subfloat[CiteSeer (Time)]{\includegraphics[width=0.23\linewidth]{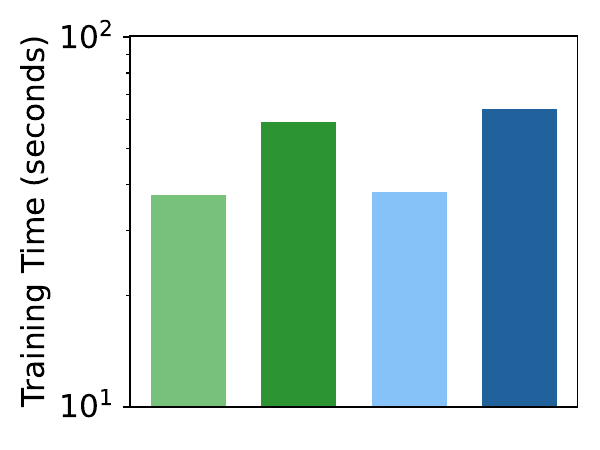}
\label{fig:cfgnn_CiteSeer-GCN_trainingtime}}
\subfloat[Amzn-Comp (Time)]{\includegraphics[width=0.23\linewidth]{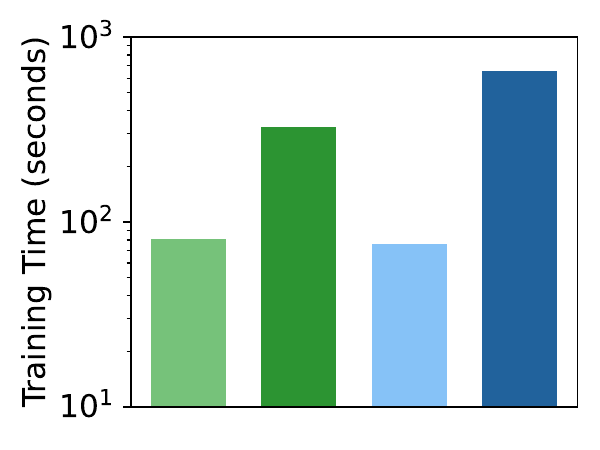}
\label{fig:cfgnn_Ammazon_Computers-GCN_trainingtime}}
\end{center}
\caption{Evaluation of different combinations of GNNs (Vanilla GCN and \name~(GCN)) and post-hoc CP methods (APS and CF-GNN) in terms of APS efficiency and training time in seconds.}
\label{fig:cfgnn}
\end{figure*}

\paragraph{Comparison with CF-GNN} In addition to the heuristic APS, we explored a parameterized CP approach CF-GNN \cite{huang2023uncertainty}, which trains a separate GNN model specifically for conformal prediction. While both CF-GNN and SparGCP study graph conformal prediction, our work focuses on optimizing GNN models during the training stage, whereas CF-GNN aims to improve the predictions of trained GNNs in the post-hoc CP procedure. These two approaches are orthogonal and may be combined. To understand their standalone performance and the effect of combining them, we conducted experiments using various baselines by picking the base classifier from vanilla GCN and \name~(using GCN as our base model) and selecting the CP method from APS and CF-GNN, respectively.

Figure \ref{fig:cfgnn} presents various combinations' efficiency and training time on two datasets. We first observe that replacing APS with CF-GNN enhances the efficiency of \name. This indicates that our work is agnostic to the downstream conformal predictor and can benefit from adopting model-based CP techniques such as CF-GNN. Additionally, \name~achieves comparable efficiency to CF-GNN; although CF-GNN is fine-tuned with each calibration set, our work does not rely on this information. Furthermore, we found the training time of our work is significantly lower (up to 76.82\%) than CF-GNN. 

\paragraph{Scaling up over OGB datasets}
To study whether \name~can scale to large datasets and effectively eliminate noisy edges from the vast number of edges in massive networks, we conducted additional experiments on Ogbn-Arxiv and Ogbn-Products, as shown in Table \ref{tbl:ogb}. Compared with vanilla GNNs, our approach consistently enhances the conformal prediction efficiency over both datasets. Specifically, \name~reduces the prediction set size by up to 11.61\% on Ogbn-Arxiv and 15.45\% on Ogbn-Products.

\begin{table*}[t]
\caption{APS efficiency improvement on OGB datasets.}
\label{tbl:ogb}
\begin{center}
\begin{tabular}{lcrrr}
\multicolumn{1}{c}{\bf Dataset} & {\bf Base Model} &\multicolumn{1}{c}{Vanilla GNN}  &\multicolumn{1}{c}{\name} &\multicolumn{1}{c}{Avg. Reduce} 
\\ \hline \\
Ogbn-Arxiv & GCN & 9.17 {\scriptsize $\pm$ 0.23} & 8.77 {\scriptsize $\pm$ 0.21} & 4.36\%\\
Ogbn-Arxiv & GAT & 10.68 {\scriptsize $\pm$ 0.35} & 9.44 {\scriptsize $\pm$ 0.36} & 11.61\%\\
Ogbn-Products & GCN & 19.41 {\scriptsize $\pm$ 1.30} & 16.26 {\scriptsize $\pm$ 0.89} & 16.23\% \\
Ogbn-Products & GAT & 21.49 {\scriptsize $\pm$ 1.93} & 17.71 {\scriptsize $\pm$ 1.23} & 17.59\% \\
\end{tabular}
\end{center}
\end{table*}

\section{Related Work}

\paragraph{Graph Conformal Prediction}
Prior research on graph conformal prediction has primarily focused on conformal node classification. Zargarbashi et al. \cite{zargarbashi2023conformal} presented DAPS, which incorporates neighborhood information into node non-conformity scores by diffusing the scores between nodes in the graph. Similarly, NAPS proposed in \cite{clarkson2023distribution} aggregates the scores from multi-hop neighbors to fully capture the structural semantics. Moreover, Huang et al. \cite{huang2023uncertainty} proposed a parameterized CP method CF-GNN that refines the non-conformity scores using a separate GNN. CF-GNN requires calibration data for training and is fine-tuned with the prediction set size on a hold-out set as its objective. While these approaches focused on developing post-hoc CP methods for node classification, our work studied enhancing CP performance in the training phase.
Recent studies have started to address other graph learning tasks. Zhao et al. \cite{zhao2024conformalized} explored the problem of conformal link prediction on graphs. Davis et al. \cite{davis2024valid} extended the validity of conformal prediction to dynamic GNNs.

\paragraph{Graph Sparsification}
Graph sparsification has been widely utilized to preserve graph structural properties and maintain learning performance. Traditional GS methods like $k$-core \cite{seidman1983kcore} generally rely on heuristics algorithms to efficiently reduce the graphs. For example, Alth{\"o}fer et al. \cite{althofer1993sparse} developed $t$-Spanner, which aims to preserve node pairwise distances in the sparsified graph. Specifically, it ensures that the distance between any two nodes in the subgraph is at most $t$ times longer than in the original graph. Satuluri et al. \cite{satuluri2011local} utilized node similarity for graph sparsification and proposed two efficient algorithms using minwise hashing. Additionally, Batson et al. \cite{batson2009twice} introduced a spectral sparsifier that constructs subgraphs with a bounded Laplacian matrix and the number of edges linear to the number of nodes. Spielman et al. \cite{spielman2011spectral} further proved that any weighted graph has a spectral sparsifier that can be generated in nearly-linear time. Motivated by the advancement of GNNs, recent studies \cite{zheng2020neuralsparse, li2023interpretable, liu2023comprehensive} have shifted to learnable graph sparsification methods. Li et al. \cite{li2023interpretable} explored the neuroscientific application of GNNs in brain graphs and trained an edge mask to eliminate task-irrelevant edges, enhancing interpretability and reducing computational overhead. Liu et al. \cite{liu2023comprehensive} extended graph sparsification by also pruning node features and GNN weights to boost training efficiency. These studies leveraged graph sparsification in traditional graph learning tasks, while our work explores the novel application of graph sparsification in enhancing conformal prediction.

\paragraph{Graph Reduction}
In addition to graph sparsification, various works have been proposed to simplify the graph structure to enhance model performance and scalability, which can be broadly categorized under the umbrella of \textit{graph reduction} \cite{hashemi2024comprehensive}. For example, Liang et al. \cite{liang2021mile} leverage \textit{graph coarsening} for scalable graph representation learning. Specifically, they first use heuristic algorithms to generate coarsened graphs by collapsing nodes, then apply an embedding algorithm to the reduced graph, and refine the embeddings using GNNs. Since graph coarsening reduces the number of nodes and edges, it significantly improves training efficiency while preserving embedding utility. Other works \cite{chen2018harp, akyildiz2020gosh} have followed similar ideas to improve scalability. More recently, motivated by data distillation, some studies have focused on \textit{graph condensation} which constructs smaller synthetic graphs to match the training trajectory with the original graph. For instance, Jin et al. \cite{jin2022graph} match the gradients between the synthetic and original graph to ensure comparable performance. Similarly, Liu et al. \cite{liu2022graph} treat the original graph as a distribution of receptive fields and construct condensed graphs by modeling and aligning the underlying distribution. These approaches have demonstrated excellent capacity for scaling GNN training over large networks. As alternatives to graph sparsification, exploring the use of graph reduction techniques in conformal prediction can be a direction for our future work.
\section{Conclusion}

In this paper, we propose \name~that leverages graph sparsification to enhance the efficiency of conformal prediction during training. \name~employs an MLP-based graph sparsifier to filter out task-irrelevant edges, and jointly trains the GNN and graph sparsifier using a CP-based objective. Through extensive experiments on various real-world datasets, we demonstrated that \name~effectively improves the efficiency of conformal prediction by reducing the size of prediction sets and consistently outperforms other baselines. Additionally, \name~adopts minibatch training that enables scaling over large networks such as OGB. Future work includes applying our approach to heterogeneous information networks \cite{hu2020heterogeneous, gurukar2022multibisage}; incorporating other CP methods \cite{zargarbashi2023conformal, clarkson2023distribution}; and extensions to other graph learning tasks such as link prediction \cite{zhao2024conformalized}.



\bibliography{iclr2025_conference}
\bibliographystyle{plain}

\appendix
\section{Details of Experimental Setup}

\subsection{Datasets}
\label{apx:datasets}

\paragraph{Dataset description} \texttt{CiteSeer} \cite{sen2008collective} and \texttt{Cora} \cite{bojchevski2018deep} are citation networks, where nodes are papers and edges denote citations. Node features are bag-of-words representations and the classification task is to predict the category of each paper. \texttt{Co-CS} (Coauthor-CS) \cite{shchur2018pitfalls} is a co-authorship graph originally from the Microsoft Academic Graph used in the KDD Cup 2016 Challenge, where each node represents an author and each edge links two authors if they have co-authored at least one paper. Keywords of research works are used as node features, and the task is to predict the author's active field of study. \texttt{Amzn-Comp} (Amazon-Computers) \cite{mcauley2015image} describes the co-purchase relationship on Amazon, where each node represents an item, and two nodes are connected if they are frequently purchased together. Here node features are bag-of-words encodings of product reviews, and the predicted label is the product category. \texttt{Ogbn-Arxiv} and \texttt{Ogbn-Products} are node classification datasets from the Open Graph Benchmark \cite{hu2020open}. Ogbn-Arxiv is a citation network of papers on Arxiv, and Ogbn-Products is a large-scale co-purchase graph. For non-OGB datasets, our experiments are conducted using the version shared by DGL \cite{wang2019dgl}. OGB datasets are from the official OGB library \footnote{\url{https://ogb.stanford.edu/}}. 

\paragraph{Data splits} For non-OGB datasets, we randomly assigned 30\%/10\% nodes to $\gV_{\mathrm{train}}, \gV_{\mathrm{valid}}$, 10\% to $\gV_{\mathrm{calib}}$, and used the remaining for test use. For OGB datasets, we adopted the predefined train/validation splits and divided the remaining nodes equally into calibration/test sets. Table \ref{tbl:data_split} shows the statistics of data splits in each dataset.

\begin{table*}[t]
\caption{Data splits of each dataset.}
\label{tbl:data_split}
\begin{center}
\begin{tabular}{lcrrrrr}
\multicolumn{1}{c}{\bf Dataset} & Split way & \multicolumn{1}{c}{\# Nodes} & \multicolumn{1}{c}{$|\gV_{\mathrm{train}}|$} & \multicolumn{1}{c}{$|\gV_{\mathrm{valid}}|$} & \multicolumn{1}{c}{$|\gV_{\mathrm{calib}}|$} & \multicolumn{1}{c}{$|\gV_{\mathrm{test}}|$}
\\ \hline \\
CiteSeer & Ratio & 3,327 & 998 & 332 & 332 & 1,665\\
Cora & Ratio & 19,793 & 5,937 & 1,979 & 1,979 & 9,898 \\
Co-CS & Ratio & 18,333 & 5,499 & 1,833 & 1,833 & 9,168 \\
Amzn-Comp & Ratio & 13,752 & 4,125 & 1,375 & 1,375 & 6,877 \\
Ogbn-Arxiv & Predefined & 169,343 & 90,941 & 29,799 & 24,301 & 24,302 \\
Ogbn-Products & Predefined & 2,449,029 & 196,615 & 39,323 & 1,106,545 & 1,106,546 \\
\end{tabular}
\end{center}
\end{table*}

\subsection{Hyperparameter choices}
\label{apx:hyperparameters}

\paragraph{GNN backbones} Our experiments adopted two GNN models, GCN \cite{kipf2017semi} and GAT \cite{kang2018self}, as standalone baselines and backbones of other approaches. Each consists of 2 GNN layers with a hidden dimensionality of 16. For GAT, we leverage the mean aggregation of two attention heads. Each model is trained for 50 epochs with a batch size of 256 and a learning rate of 0.01. For OGB datasets, we increase the batch size to 512 and the dimensionality of hidden representations to 128. On Ogbn-Products, we employed a neighbor sampler with fanouts of [10, 10] for message passing, while on other datasets we used the full neighborhood information.

Additionally, other baselines have their own specific hyperparameters. We explored a wide range of choices and reported the optimal performance. Specifically, we tuned each baseline as follows:

\paragraph{$k$-core} The key hyperparameter of $k$-core is $k$, which is the minimum degree of nodes in the generated subgraph. To fully explore the performance with different numbers of edges removed, we selected $k$ from $[1, 2, 3, 4, 5]$ on CiteSeer, from $[2, 4, 6, 8, 10]$ on Cora and Co-CS, and from $[2, 4, 6, 8, 10, 12, 16, 20, 24, 28, 32, 36, 40, 44, 48, 52]$ on Amzn-Comp. 

\paragraph{DropEdge} We tuned the probability $p$ of dropping an edge in the message flow graph by iterating over $[0.1, 0.2, 0.3, 0.4, 0.5, 0.6, 0.7, 0.8, 0.9, 0.99]$ across all datasets.

\paragraph{PTDNet} PTDNet utilized two regularizations to train their graph denoising model. Therefore we tried different configurations of corresponding weights in their objective. Specifically, we select the value of $\beta_{1}$ (for non-zero entries in the sparsified adjacency matrix) from $[0, 0.05, 0.1, 0.5, 1]$ and the value of $\beta_{2}$ (for low-rank regularizer $\gR_{lr}$) from $[0, 0.05, 0.1]$. These values were also used in their experiments.

\paragraph{NeuralSparse} We varied two hyperparameters of NeuralSparse: the number of sampled edges for each node $k$ and the dimensionality of hidden channels $d$ in their graph sparsifier. Specifically, $k$ is chosen from $[2, 4, 8, 16]$ and $d$ is from $[16, 32, 64]$.

\paragraph{CF-GNN} We chose GAT as the base architecture of CF-GNN and tuned it with the following hyperparameters: (1) learning rate from $\mathrm{LogUniform}[10^{-4}, 10^{-1}]$, (2) GNN hidden dimension from $\{16, 32, 64, 128\}$, (3) number of GNN layers from $[1, 2, 3, 4]$, (4) attention heads from $[2, 4, 8]$, and (5) temperature factor $\tau$ from $\mathrm{LogUniform}[10^{-3}, 10^{1}]$.

\paragraph{\name~(Our work)} In this study, we tuned the following hyperparameters in our approach: (1) the strength of edge dropping $\gamma$ from $[0.1, 0.2, 0.3, 0.4, 0.5, 0.6, 0.7, 0.8, 0.9, 0.99]$; (2) the weight of our CP-based loss $\lambda$ from $\mathrm{Uniform}[0.1, 1]$ and $\mathrm{Uniform}[1, 10]$. To align with the miscoverage rate in the downstream CP task, we let $\alpha_{\mathrm{train}}$ in our CP-based loss be equal to $\alpha=0.1$. The hidden dimensionality of the graph sparsifier is set to match the base GNN dimensionality.

\subsection{Implementation}

We implemented vanilla GNNs, $k$-core, DropEdge, and our approach in PyTorch and DGL. For PTDNet and NeuralSparse, we use the implementation from Graph Structure Learning Benchmark (GSLB) \cite{li2023gslb}. Our code and data will be available upon acceptance.
\section{Impact of Sparsification Strength}
\label{apx:sparsification_strength_full}
\begin{figure*}[t]
\begin{center}

\includegraphics[width=0.6\linewidth]{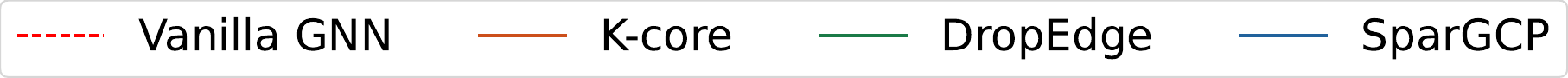}\\


\subfloat[CiteSeer, GAT]{\includegraphics[width=0.24\linewidth]{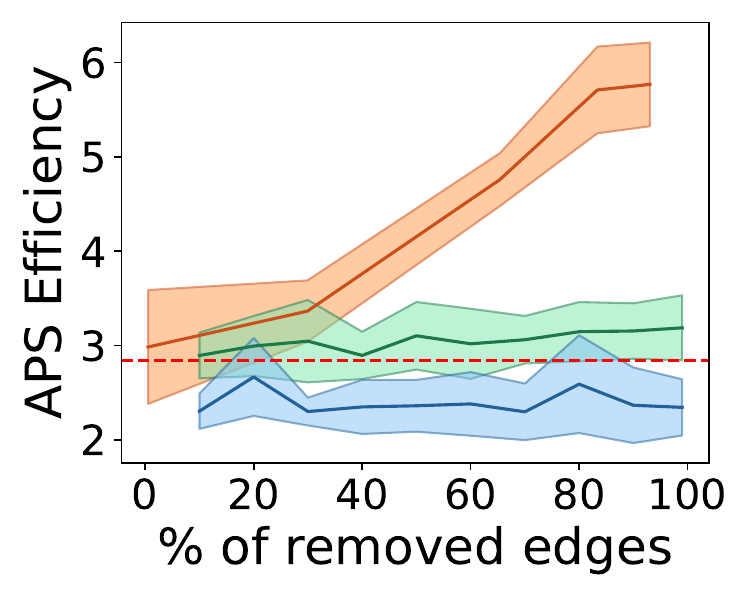}
\label{fig:reducing_edges_full_GAT_CiteSeer}}
\subfloat[Cora, GAT]{\includegraphics[width=0.24\linewidth]{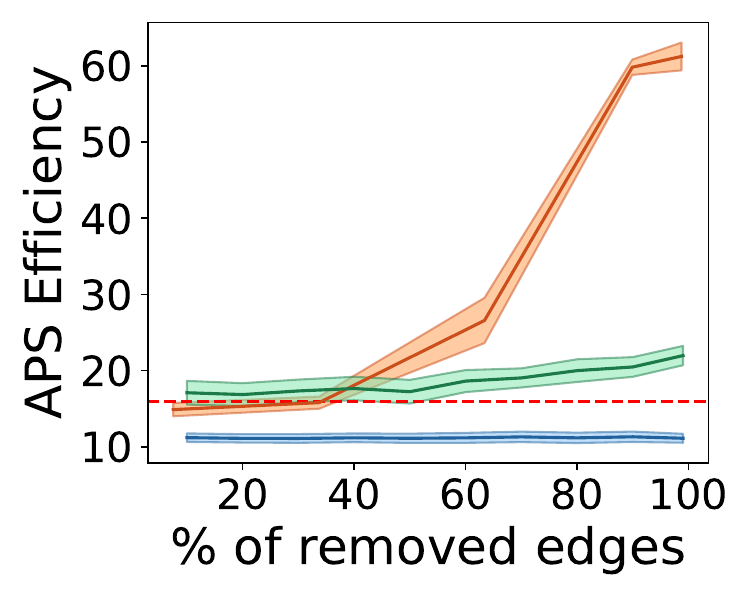}
\label{fig:reducing_edges_full_GAT_Cora}}
\subfloat[Co-CS, GAT]{\includegraphics[width=0.24\linewidth]{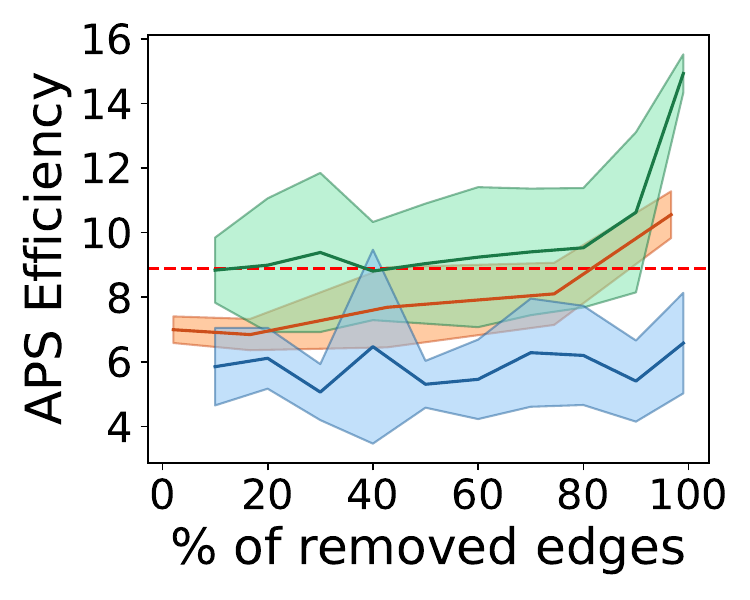}
\label{fig:reducing_edges_full_GAT_Coauthor_CS}}
\subfloat[Amzn-Comp, GAT]{\includegraphics[width=0.24\linewidth]{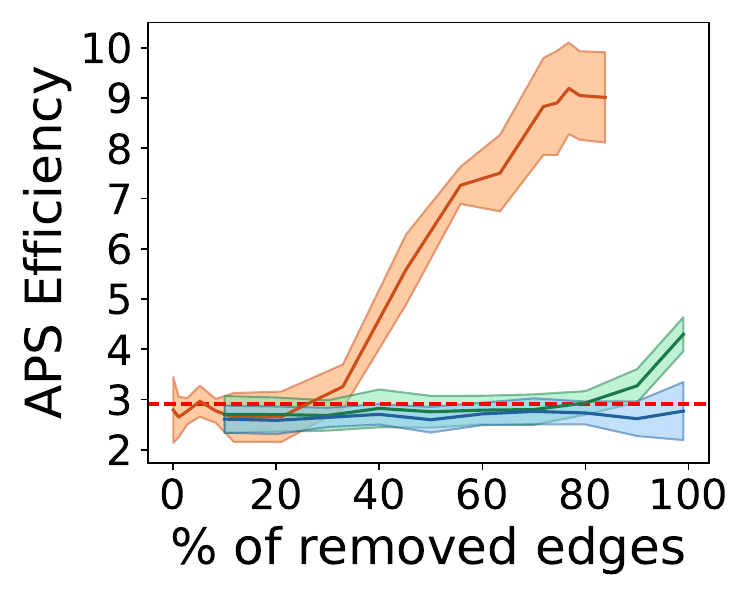}
\label{fig:reducing_edges_full_GAT_Amazon_Computers}}
\end{center}
\caption{APS efficiency of baselines using GAT with varying strength of graph sparsification.}
\label{fig:removing_edges_full}
\end{figure*}

Figure \ref{fig:removing_edges_full} presents the results of APS efficiency v.s. the strength of graph sparsification across four datasets using GAT as the base GNN model. Consistent with the findings in the main paper, \name~remains effective as more edges are removed, whereas the performance of other baselines significantly declines when a large number of edges are filtered.

\begin{figure*}[t]
\begin{center}


\subfloat[CiteSeer, GAT]{\includegraphics[width=0.24\linewidth]{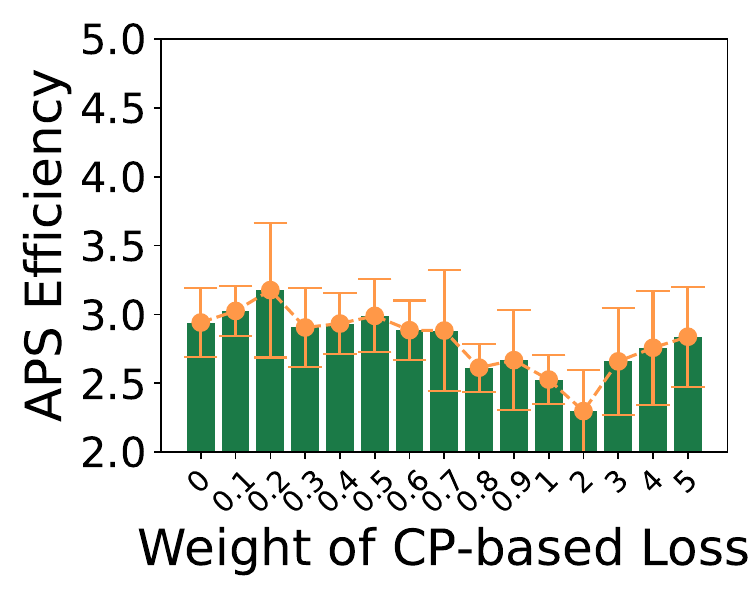}
\label{fig:varying_weight_full_GAT_CiteSeer}}
\subfloat[Cora, GAT]{\includegraphics[width=0.24\linewidth]{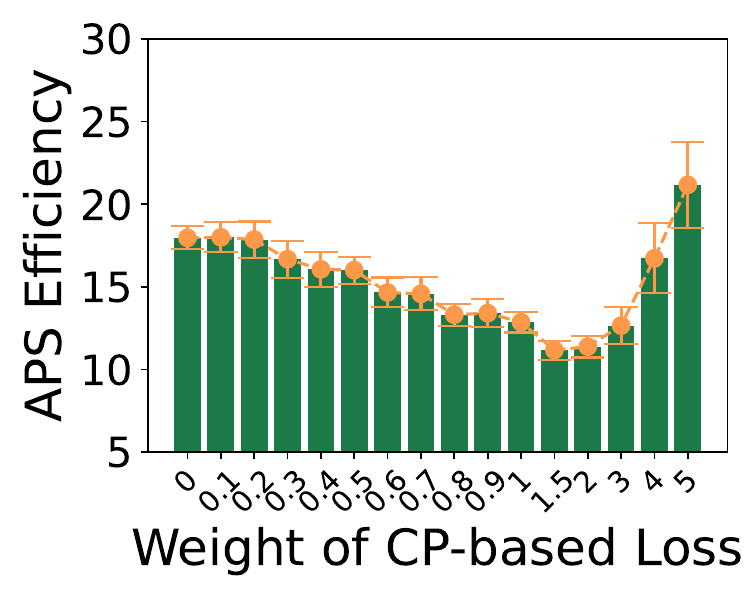}
\label{fig:varying_weight_full_GAT_Cora}}
\subfloat[Co-CS, GAT]{\includegraphics[width=0.24\linewidth]{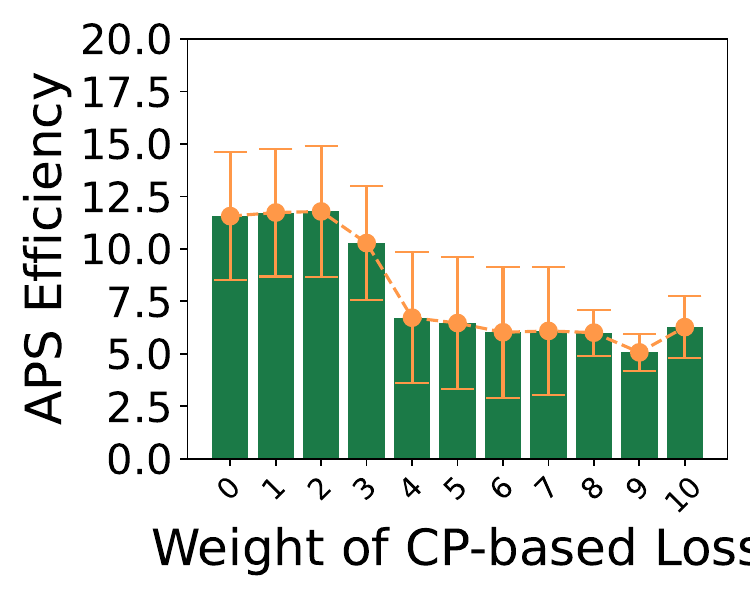}
\label{fig:varying_weight_full_GAT_Coauthor_CS}}
\subfloat[Amzn-Comp, GAT]{\includegraphics[width=0.24\linewidth]{figs/varying_weight/VaryingWeight_Amazon_Computers-GCN.pdf}
\label{fig:varying_weight_full_GAT_Amazon_Computers}}
\end{center}
\caption{APS efficiency of baselines using GAT with varying the weight of CP-based objective.}
\label{fig:varying_weight_full}
\end{figure*}

\section{Varying the Weight of CP-based Loss}
\label{apx:varying_weight_full}

Figure \ref{fig:varying_weight_full} shows the complete results of the impact of varying the CP-based objective weight on the CP efficiency of GAT-based baselines. The findings from the main paper are consistent across additional datasets and GNN backbones, further validating the trends observed in our approach.

\end{document}